\documentclass{article}
\usepackage{xcolor}
\usepackage[preprint]{corl_2026}
\usepackage[T1]{fontenc}
\usepackage[utf8]{inputenc}
\usepackage{graphicx}
\usepackage{pdfpages}
\usepackage{amsmath,amssymb}
\usepackage{booktabs}
\usepackage{multirow}
\usepackage{tabularx}
\usepackage{array}
\usepackage{placeins}
\usepackage{microtype}
\usepackage[font=small,labelfont=bf]{caption}
\newcolumntype{Y}{>{\raggedright\arraybackslash}X}
\graphicspath{{figures/}}

\title{Verified experience improves materials prediction with large language models}
\author{
Yiming Zhang$^1$ \qquad Koji Tsuda$^{1,2,3}$\\
$^1$Graduate School of Frontier Sciences, The University of Tokyo, Chiba, Japan\\
$^2$Center for Basic Research on Materials,\\
National Institute for Materials Science, Tsukuba, Ibaraki, Japan\\
$^3$RIKEN Center for Advanced Intelligence Project, Tokyo, Japan\\
\texttt{tsuda@k.u-tokyo.ac.jp}
}
\hypersetup{
pdftitle={Verified experience improves materials prediction with large language models},
pdfauthor={Yiming Zhang and Koji Tsuda},
pdfsubject={Materials prediction with verified episodic memory},
pdfkeywords={materials prediction, episodic memory, sequential benchmark}
}
\begin{document}
\maketitle

\begin{abstract}
Materials discovery is an iterative loop: experiments and calculations verify predictions, and each outcome can inform the next. Yet most predictors are specialized single-task models fixed after training. Large language models have recently emerged as predictors across heterogeneous tasks but generally underperform specialized models, and retrieval-augmented variants discard their own outcomes. Benchmarks also score queries in isolation, leaving improvement from outcomes untested. Here we construct MatMemBench, a sequential benchmark of 22 materials tasks that reveals outcomes only after prediction and tests online adaptation, memory retention and cross-task transfer. We present MatRecall, a framework that lets large language models store verified predictions, with evidence, outcome and error, as episodes and recall them without retraining. MatRecall improved median performance by 62\% over a memory-free model, exceeded specialized models on 11 tasks, remained useful after feedback stopped and transferred between related tasks. Verified experience could become a reusable resource for self-improving discovery loops.
\end{abstract}
\keywords{Materials prediction, Episodic memory, Sequential evaluation}

\section{Introduction}

Machine-learning models have become central to materials prediction. Crystal graph networks and periodic transformers exploit atomic geometry to predict properties with high accuracy \citep{Xie2018CGCNN,Chen2019MEGNet,Choudhary2021ALIGNN,Yan2022Matformer}, and shared resources such as JARVIS and Matbench provide standardized data and comparisons \citep{Choudhary2020JARVIS,Dunn2020Matbench}. Beyond properties, specialized models predict synthesis precursors, process conditions, action graphs and experimental outcomes \citep{Raccuglia2016Failed,Kim2020SynthesisPlanning,He2023Precursor,Prein2025DataAugmented,Yang2022PcMSP,Akbari2022MeltpoolNet,Malik2021SynthesisOutcomes}. This accuracy, however, comes from specialization. Each model is designed, trained and tuned for one input--output mapping and remains fixed after training. A new task therefore requires a new model, and outcomes observed after deployment do not inform later predictions unless the model is retrained. Specialized predictors thus provide no common mechanism for accumulating experience across the heterogeneous tasks that make up a materials campaign.

Large language models (LLMs) offer a complementary route. A single model can accept compositions, structures, synthesis text, process variables and requested output schemas without a task-specific architecture. This generality, however, comes at the cost of numerical precision. Systematic experiments with MatText showed that LLMs learn categorical patterns more readily than coordinate relations, leaving a persistent gap from geometry-aware models on structure-dominated tasks \citep{Alampara2026MatText}. Retrieval-augmented generation (RAG) partly mitigates this weakness by grounding predictions in training records \citep{Lewis2020RAG,Jiang2026PIRAG}, and tool-using agents can further decide when to retrieve, calculate or revise \citep{Yao2023ReAct,Bran2024ChemCrow,Boiko2023Coscientist}. Materials systems also increasingly preserve synthesis provenance as structured graphs, creating a basis for evidence-grounded process reasoning \citep{Tsuruta2025MatPROV,Zhang2026ProvMind}. Most materials RAG systems, however, retrieve from a static corpus. The model sees previous examples, but its own completed predictions, and the errors they reveal, are discarded rather than retained for similar queries.

This limitation matters because materials research is inherently sequential. In a screening or optimization campaign, each prediction is followed by a calculation or experiment, and its verified outcome becomes available before the next decision. We therefore formulate materials prediction as a sequential task. Queries arrive one at a time, each answer is revealed only after the corresponding prediction has been recorded, and the predictor may use earlier outcomes but never the current one. Under this formulation, every prediction defines a natural event: it begins with a query, proceeds through retrieval and tool use, and ends when a computed or measured outcome verifies the prediction. The central question then shifts from how accurately a model predicts in isolation to whether it improves as such events accumulate. Materials tasks make this question measurable. Related materials and processes recur within a campaign, physically related properties link different tasks, and outputs range from numerical properties to precursor sets and operation sequences. Standard materials benchmarks, however, score each test query independently of earlier outcomes \citep{Dunn2020Matbench,Alampara2026MatText}.

Episodic memory provides a principled way to retain these events. Cognitive science distinguishes general factual knowledge from memories of individual events and their context. Building on this distinction, a recent framework for long-term LLM agents operationalizes episodic memory through five properties: long-term storage, explicit reasoning, single-shot learning, instance specificity and contextual relations \citep{Pink2025EpisodicMemory}. In this view, a static training library plays a semantic-like role by supplying stable domain knowledge, whereas episodic memory records what happened during a particular interaction and reinstates it in the model's working context when a related situation arises. A verified prediction event fits this definition closely. Encoding can bind the query, actions, evidence and error into one contextualized record, and later retrieval can reinstate that record without updating model parameters. Studies of LLM agents have already used stored experiences, synthesized reflections and verbal feedback to guide subsequent decisions \citep{Park2023GenerativeAgents,Shinn2023Reflexion,Zhao2024ExpeL}. It remains unknown, however, whether such episodes improve materials prediction beyond static evidence, remain useful after feedback stops and transfer between tasks.

\begin{figure}[!htbp]
\centering
\includegraphics[width=\textwidth]{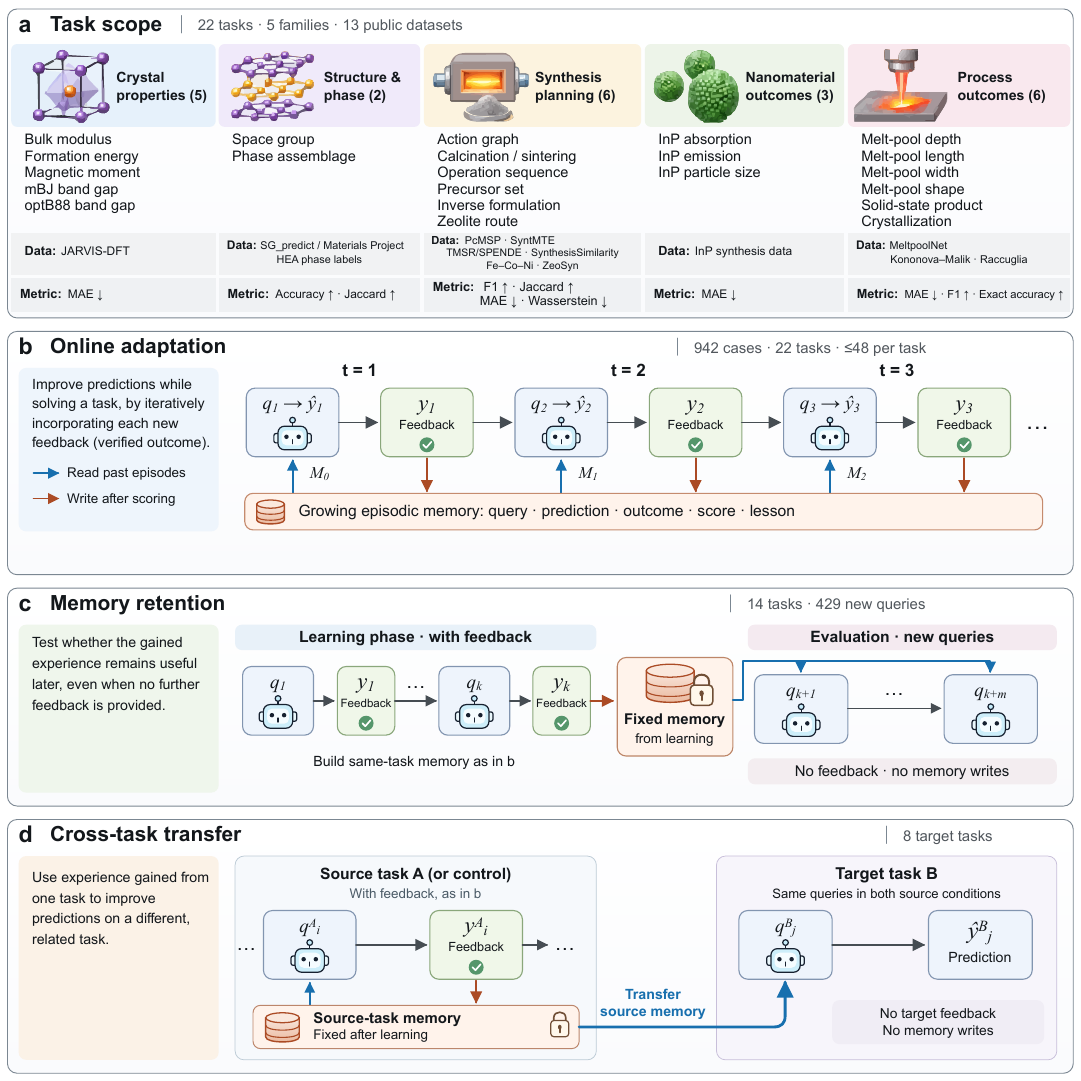}
\caption{\textbf{MatMemBench evaluates the accumulation, retention and cross-task reuse of verified experience.}
\textbf{a}, Task scope: 22 tasks from 13 public datasets span crystal properties, structure and phase, synthesis planning, nanomaterial outcomes and process outcomes. Colours distinguish task families; arrows beside native metrics indicate the favourable direction.
\textbf{b}, Online adaptation evaluates 942 cases across 22 tasks, with at most 48 cases per task. The illustrated episodic-memory predictor uses $\mathcal{M}_{t-1}$ to predict $\hat{y}_t$ for query $q_t$. Outcome $y_t$ is then revealed and scored ($\ell_t$), and the completed episode is written to memory. Episodic memory starts empty ($\mathcal{M}_0=\varnothing$); memory use depends on the method. A fixed seed and case identifiers determine the default order; a separate 654-case cohort tests five shuffled and five group-blocked orders.
\textbf{c}, Memory retention separates learning with feedback from evaluation without feedback. After $k$ learning queries, the same-task memory remains fixed while predicting 429 new, group-disjoint queries across 14 tasks.
\textbf{d}, Cross-task transfer applies memory acquired on source task A to target task B. Each of eight targets compares memories from a relevant source and an unrelated control on identical target queries. In \textbf{c} and \textbf{d}, evaluation provides no target feedback or memory writes. Robot boxes denote LLM prediction, blue arrows read memory, terracotta arrows write after scoring, and locks denote fixed memories.}
\label{fig:benchmark}
\end{figure}

Here we construct MatMemBench to measure whether materials predictors improve from verified outcomes, and introduce MatRecall to retain and reuse those outcomes. MatMemBench contains 22 tasks from 13 public datasets across five materials task families, each scored in its native metric (Fig.~\ref{fig:benchmark}a). Each task uses a reproducible default query order defined by a fixed sampling seed and case identifiers. Three evaluation settings distinguish how experience is acquired and reused. Online adaptation reveals each outcome after prediction and allows experience to accumulate within a task (Fig.~\ref{fig:benchmark}b). Memory retention applies the acquired same-task memory to new queries after feedback stops (Fig.~\ref{fig:benchmark}c). Cross-task transfer compares fixed memories from a relevant source task and an unrelated control on identical target queries (Fig.~\ref{fig:benchmark}d). Retention and transfer therefore test two uses of experience acquired online: reuse within a task and reuse across tasks.

\begin{figure}[p]
\centering
\includegraphics[width=\textwidth]{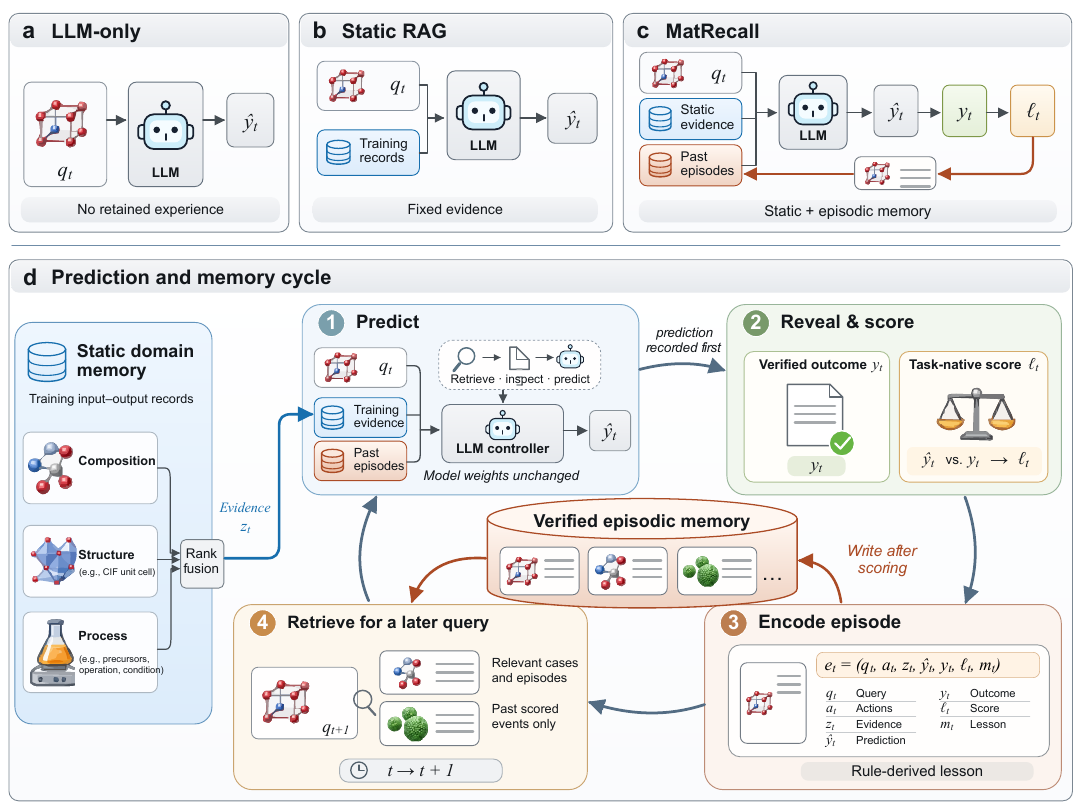}
\caption{\textbf{MatRecall combines static domain evidence with verified episodic memory for materials prediction.}
\textbf{a}, LLM-only predicts each query without retaining experience.
\textbf{b}, Static RAG retrieves training input--output records from a fixed domain library.
\textbf{c}, MatRecall retrieves both static evidence and past episodes, then stores the completed prediction event after outcome verification and scoring.
\textbf{d}, The four-stage memory cycle. First, the LLM controller retrieves evidence, inspects prior experience and records a prediction. Composition, structure and process retrieval views supply ranked training records, which are fused into the evidence context. Second, the outcome is revealed and compared with the prediction using the task-native score. Third, the completed event is encoded as $e_t=(q_t,a_t,z_t,\hat{y}_t,y_t,\ell_t,m_t)$: query, actions, retrieved evidence, prediction, verified outcome, score and a rule-derived lesson, respectively. Fourth, relevant cases and episodes are retrieved for a later query. Blue denotes static training evidence and terracotta denotes episodic memory; the central store receives writes after scoring and supplies reads for later predictions. Model parameters remain unchanged throughout the cycle.}
\label{fig:overview}
\end{figure}

MatRecall combines fixed training evidence with a growing record of verified prediction events (Fig.~\ref{fig:overview}). LLM-only handles each query independently, whereas static retrieval supplies training examples without retaining completed predictions (Fig.~\ref{fig:overview}a,b). MatRecall adds episodic memory to the same prediction process (Fig.~\ref{fig:overview}c). Its four-stage cycle predicts, reveals and scores the outcome, encodes the completed episode, and retrieves relevant experience for a later query (Fig.~\ref{fig:overview}d). Static domain memory indexes training input--output records through composition, structure and process views. Each episode binds the query, actions, retrieved evidence, prediction, outcome, score and compressed lesson. The LLM uses both memory types in its working context while its parameters remain unchanged.

On MatMemBench, MatRecall improved median task performance during online adaptation by 61.7\% over the memory-free LLM-only model and 14.5\% over static retrieval, with the largest gains after related experience had accumulated. After feedback stopped, the acquired memory still outperformed static retrieval on all five crystal-property tasks. In cross-task transfer, memories from relevant source tasks reduced error by 13.0--54.6\% relative to unrelated controls when combined with static evidence. MatRecall also narrowed the numerical gap between the LLM and specialized models, closing 49--99\% of it on the five crystal-property tasks, and exceeded specialized models on 11 of the 22 tasks, mostly in synthesis and process prediction. More broadly, these results suggest that the verified experience of an LLM can be accumulated, retained and reused much as data and models are, pointing towards discovery systems that become more reliable with each calculation or experiment they inform.

\FloatBarrier

\section{Results}

\subsection{Verified prediction episodes improve online adaptation}

We first asked whether feedback from earlier predictions improves later predictions within the same task. The online-adaptation setting contains 942 sequentially presented cases from 22 tasks, with at most 48 cases per task (Fig.~\ref{fig:benchmark}a,b). Each task retains its native metric and unit. At step $t$, every LLM method predicts the same query before the target is revealed; methods differ only in which information from steps $1$ to $t-1$ remains available. The task-specific memory acquired by the end of each sequence supplies the later retention and transfer evaluations.

Six methods separate static retrieval from experience accumulation. LLM-only is stateless. Static RAG retrieves from a fixed library using the views appropriate for the task. ReAct + static RAG can choose retrieval views and tools but cannot encode experience. Rolling-context in-context learning (ICL) places the most recent verified cases directly in the prompt. Dynamic case RAG writes each query--outcome pair to a retrievable case store. MatRecall encodes the complete verified episode and allows the controller to retrieve static evidence, prior cases and trajectory-level lessons. The LLM, output schema and visible task input are held constant across methods.

\begin{figure}[!tbp]
\centering
\includegraphics[width=\textwidth]{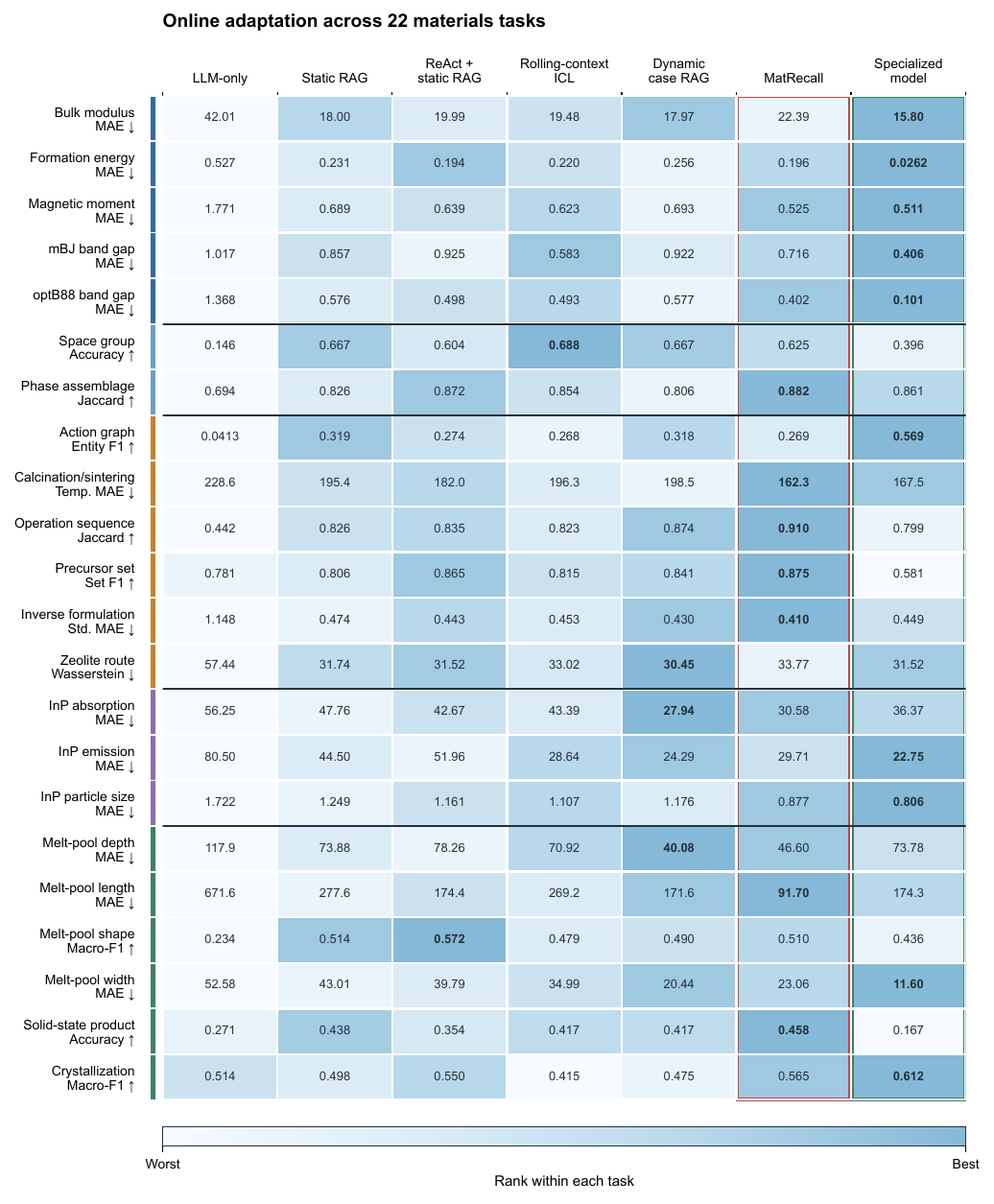}
\caption{\textbf{Verified episodic memory improves online adaptation across 22 materials tasks.}
Each row is one task and each cell reports the task-native numerical value over the same 942-case online-adaptation sequence. Arrows beside metric names indicate the favourable direction. Colours show the rank of each method within its row, from worst to best, so a single extreme value does not compress the contrast among the others; colours are comparable within a task but not across tasks. Values that print identically share a rank. Bold type identifies the best method, coloured side bars denote task families, and red and green outlines identify MatRecall and the specialized model, respectively. The six LLM methods are evaluated sequentially with their stated memory condition; the specialized model predicts the same queries without online updates.}
\label{fig:sequential}
\end{figure}

Across the 22 tasks, MatRecall improved median task performance by 61.7\% over LLM-only, 14.5\% over static RAG, 10.0\% over ReAct + static RAG, 10.3\% over rolling-context ICL and 4.5\% over dynamic case RAG (Fig.~\ref{fig:sequential}). It outperformed LLM-only on all 22 tasks, static RAG on 17 and dynamic case RAG on 14. The gains extended across numerical, set-valued and sequential outputs. Relative to LLM-only, MatRecall reduced formation-energy MAE by 62.9\% (0.527 to 0.196), InP particle-size MAE by 49.1\% (1.72 to 0.877~nm) and melt-pool-length MAE by 86.3\% (672 to 91.7~$\mu$m). Operation-sequence Jaccard increased 2.06-fold, from 0.442 to 0.910. Relative to static RAG, the corresponding changes were a 15.5\% reduction in formation-energy MAE, a 29.8\% reduction in particle-size MAE, a 67.0\% reduction in melt-pool-length MAE and a 10.2\% increase in operation-sequence Jaccard. Static RAG remained better for bulk modulus, space group, action-graph extraction, zeolite route distribution and melt-pool shape, and dynamic case RAG achieved the best LLM score on six tasks, including four InP and melt-pool regression tasks. Figure~\ref{fig:sequential} reports every score on its original task-native scale.

The specialized models predicted the same queries without online updates (Fig.~\ref{fig:sequential}). LLM-only exceeded the specialized model on 2 of the 22 tasks and MatRecall on 11, mostly in synthesis, nanomaterial and process prediction. Across the five crystal-property tasks, MatRecall closed 49.3--98.9\% of the gap between LLM-only and the specialized model, and its error was 1.03--7.46 times the specialized-model error. MatRecall was stronger on several sequence and process outcomes: operation-sequence Jaccard was 0.910 versus 0.799, melt-pool-length MAE was 91.7 versus 174~$\mu$m and InP absorption MAE was 30.6 versus 36.4~nm. The value of episodic feedback therefore depends on whether prior outcomes or task-specific inductive bias better match the prediction target.

\subsection{Verified case memory supports robust online adaptation}

To isolate the contributions of feedback content, query order and accumulated experience, we ran a separate online-adaptation study on the same 22 tasks with 11 memory variants and ten query sequences. Five sequences randomly shuffle the same cases, whereas five place cases from the same material or process group next to one another. Each sequence contains at most 32 cases per task and 654 cases in total. The experience analysis separates the first query in each group from later queries in the 15 tasks that contain repeated group members. Retrieval-budget sensitivity is reported in Supplementary Fig.~S1.

\begin{figure}[!tbp]
\centering
\includegraphics[width=\textwidth]{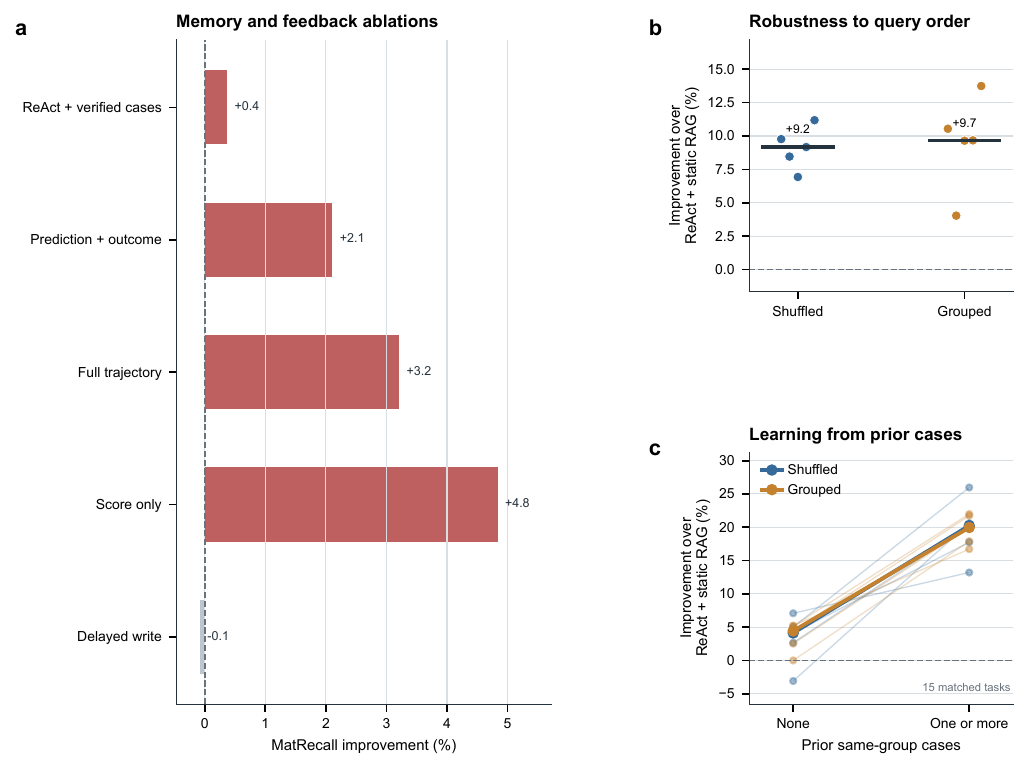}
\caption{\textbf{Verified case memory supports online adaptation across query orders and accumulated experience.}
\textbf{a}, Median improvement of MatRecall over five memory and feedback ablations on the 654-case reference sequence. Bars show task-level median changes, and positive values favour MatRecall. Comparisons with the baseline methods are shown for the online-adaptation sequence in Fig.~\ref{fig:sequential}.
\textbf{b}, Improvement over ReAct + static RAG for five shuffled and five grouped query sequences. Points denote individual sequences and horizontal lines denote the median across the five sequences.
\textbf{c}, Improvement over ReAct + static RAG for queries with no prior case from the same material or process group and for queries with one or more prior cases. Thin lines denote individual sequences and thick lines denote the median across five sequences. This comparison uses the same 15 tasks in both experience conditions. Dashed lines denote no improvement.}
\label{fig:mechanisms}
\end{figure}

Among the memory and feedback ablations on the reference sequence (Fig.~\ref{fig:mechanisms}a), delaying memory writes changed the median by $-0.1\%$. Verified outcomes, rather than immediate write timing, drove the aggregate gain.

Full trajectories and verified case memory produced similar aggregate performance. MatRecall differed by 3.2\% from the full-trajectory ablation and by 0.4\% from ReAct + verified cases, which stores verified cases without trajectory fields. Across the five shuffled sequences, its median improvement over ReAct + static RAG ranged from 6.9\% to 11.2\%, with a median of 9.2\%. Across the five grouped sequences, the corresponding range was 4.0\% to 13.7\%, with a median of 9.7\% (Fig.~\ref{fig:mechanisms}b).

Accumulated same-group experience produced a larger separation. In the 15 tasks containing repeated groups, the median improvement over ReAct + static RAG increased from 4.1\% with no prior same-group case to 20.2\% with one or more prior cases under shuffled order. Under grouped order, it increased from 4.4\% to 20.0\% (Fig.~\ref{fig:mechanisms}c). Thus, the larger gains occurred after relevant cases became available, whether groups were interleaved or contiguous.

\subsection{Stored experience remains useful after feedback stops}

The memory-retention setting uses the experience acquired during the 942-case online-adaptation sequence (Fig.~\ref{fig:benchmark}b,c). The primary cohort contains 14 tasks and 429 new queries separated from online-adaptation cases by source-native material, document, topology or process groups. Every query used the same fixed task-specific memory. Targets remained hidden during evaluation and no new episode was written.

\begin{figure}[!tbp]
\centering
\includegraphics[width=\textwidth]{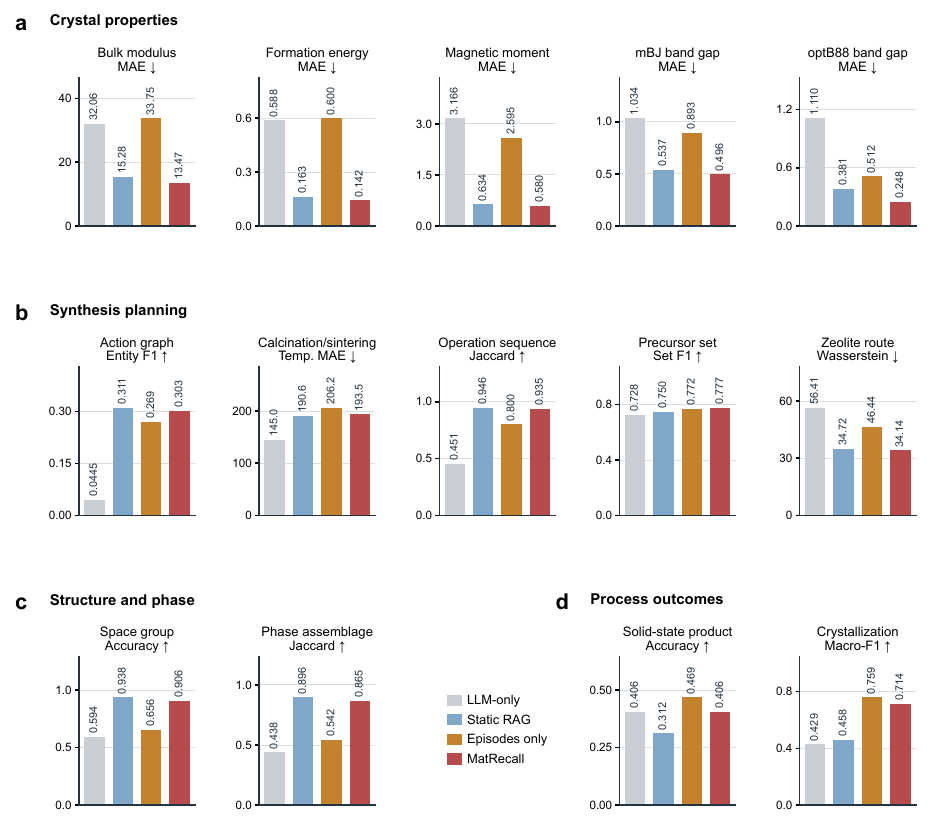}
\caption{\textbf{Verified experience remains useful after feedback stops.}
Task-native values for 429 new same-task queries from 14 tasks, drawn from groups absent during online adaptation. The acquired memory is fixed, answers are not revealed and no method writes memory. \textbf{a}, Crystal properties. \textbf{b}, Synthesis planning. \textbf{c}, Structure and phase. \textbf{d}, Process outcomes. Each small panel shows one task on its own axis, and arrows indicate the favourable direction. Episodes only uses verified episodes without static evidence; MatRecall combines static evidence with episodes. Values for cases only, dynamic case RAG, ReAct + static RAG and the specialized models are given in Supplementary Table~S6.}
\label{fig:retention}
\end{figure}

With the acquired memory fixed, MatRecall improved on LLM-only in 12 of the 14 tasks (Fig.~\ref{fig:retention}); it matched LLM-only for solid-state product prediction and was worse for calcination temperature (193.5 versus 145.0~$^{\circ}$C). Formation-energy MAE decreased from 0.588 to 0.142 and bulk-modulus MAE from 32.1 to 13.5~GPa. Much of this improvement, however, was already provided by static retrieval, which reduced the same errors to 0.163 and 15.3~GPa (Supplementary Table~S6). We therefore assessed retained experience against static RAG. MatRecall outperformed static RAG in nine tasks, including all five crystal-property tasks, where MAE decreased by 7.6--34.8\%. Beyond crystal properties, retained experience increased crystallization macro-F1 from 0.458 to 0.714 and solid-state exact-formula accuracy from 0.313 to 0.406. Static RAG remained slightly better for space group, phase assemblage, action-graph extraction and operation sequence, with differences of at most 0.031 in score, and for calcination temperature, with a difference of 2.9~$^{\circ}$C. Experience acquired online therefore retained predictive value after feedback stopped, but its benefit beyond static evidence depended on the task.

Static grounding determined whether episodes helped structure-sensitive regression (Fig.~\ref{fig:retention}). For bulk modulus, MAE was 32.1~GPa for LLM-only and 33.8~GPa for episodes only, but decreased to 13.5~GPa when the same episodes were combined with static structural evidence. Formation-energy MAE followed the same pattern: 0.588, 0.600 and 0.142. Verified experience supplies calibration and strategy information, while structure-dependent prediction also requires a grounded material analogue.

Specialized models retained a large advantage for crystal properties. Across the five crystal-property tasks, MatRecall closed 60.3--84.7\% of the performance gap between LLM-only and the specialized model. Its errors remained 3.1, 5.8 and 3.5 times the specialized-model errors for bulk modulus, formation energy and the mBJ band gap, respectively (Supplementary Table~S6). For operation-sequence prediction, MatRecall reached a Jaccard score of 0.935 compared with 0.892 for the specialized model; crystallization macro-F1 was 0.714 and 0.608, respectively. Episodic memory narrows the gap when comparable cases and textual or categorical context are informative, while geometric inductive bias remains decisive for structure-sensitive targets.

We next tested this boundary on the independent MatText benchmark, using three property tasks and five aligned crystal-text representations \citep{Alampara2026MatText}. MatRecall reduced MAE relative to LLM-only in all 15 task--representation comparisons (Supplementary Table~S8). In a controlled coordinate--category benchmark, MatRecall reduced MAE from 0.650 to 0.238 when the target depended entirely on coordinates. The specialized geometry model reached an MAE of 0.0236 on the same queries (Supplementary Table~S9). Verified experience improved numerical calibration across representations, while geometry-aware features remained the strongest basis for coordinate-dominated prediction.

\subsection{Relevant source memories improve cross-task transfer}

Cross-task transfer requires source experiences that are relevant to the target task. The cross-task transfer setting pairs each of eight targets with one relevant source task and one unrelated control task while keeping the target queries fixed (Fig.~\ref{fig:benchmark}d). Four primary pairs share property-regression schemas: formation energy to bulk modulus, bulk modulus to formation energy, optB88 band gap to mBJ band gap and mBJ band gap to optB88 band gap. Four additional pairs connect phase, symmetry or synthesis tasks. Source relevance was assigned from task semantics and output schema before evaluation.

\begin{figure}[!tbp]
\centering
\includegraphics[width=\textwidth]{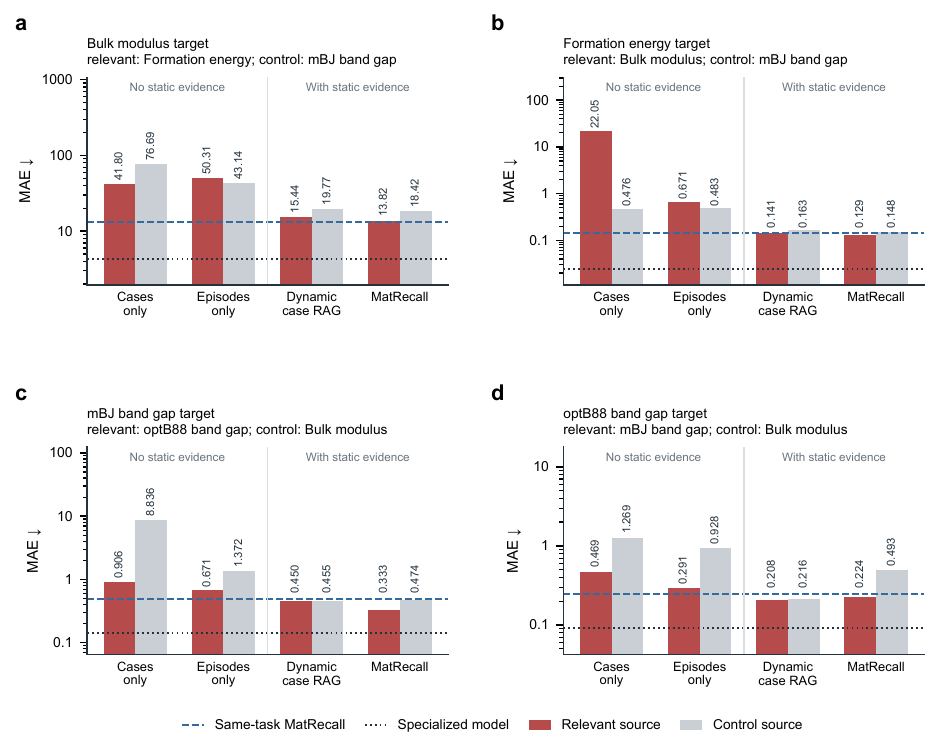}
\caption{\textbf{Relevant source memories improve cross-task transfer when combined with static evidence.}
\textbf{a}--\textbf{d}, MAE on four property targets with fixed memory from a relevant source task (red) or an unrelated control task (grey), evaluated on identical target queries. Cases only and episodes only omit static evidence; dynamic case RAG and MatRecall include static grounding. Dashed and dotted lines mark same-task MatRecall and the specialized model on the same queries. The vertical axis is logarithmic. Values for all eight targets are given in Supplementary Table~S7.}
\label{fig:cross}
\end{figure}

With static grounding, relevant source episodes reduced MAE by 25.0\% for bulk modulus, 13.0\% for formation energy, 29.7\% for the mBJ band gap and 54.6\% for the optB88 band gap relative to the unrelated controls (Fig.~\ref{fig:cross}). Dynamic case RAG showed the same direction. Because each relevant--control comparison uses identical target queries, these changes measure the effect of source-task memory selection.

Transfer without static evidence produced heterogeneous effects. Formation-energy episodes increased bulk-modulus error without structural grounding, while bulk-modulus cases produced large formation-energy errors when used alone. Cross-schema transfer was neutral or detrimental for most additional pairs (Supplementary Fig.~S3). Static domain evidence anchors cross-task transfer and limits source memories that do not preserve the target's input--output relation.

\FloatBarrier

\section{Discussion}

Our results show that a general-purpose LLM can improve during a materials campaign without retraining when each verified prediction is retained as retrievable experience. Treating predictions as scored events rather than isolated queries allowed MatRecall to outperform static retrieval on 17 of the 22 MatMemBench tasks during online adaptation and to exceed the specialized models on 11 of them, compared with 2 for the memory-free model. Verified experience therefore narrows, and on several synthesis and process tasks closes, the gap that usually separates general LLMs from task-specific predictors.

Three observations indicate how this experience becomes useful. First, its value grows with relevance: the median improvement over ReAct + static RAG rose from about 4\% for queries without a prior case from the same material or process group to about 20\% once such a case was available, under both shuffled and grouped query orders. Second, episodic and static memory are complementary rather than interchangeable. After feedback stopped, episodes alone did not improve bulk-modulus or formation-energy prediction, whereas the same episodes combined with static structural evidence reduced bulk-modulus MAE from 32.1 to 13.5~GPa. In the terms of the episodic-memory framework that motivated this work \citep{Pink2025EpisodicMemory}, the static library supplies a semantic-like anchor for the material at hand, and verified episodes supply the calibration and strategy for using it. Third, the verified outcome carried most of the information: MatRecall differed by only 0.4\% from ReAct + verified cases, which stores outcomes without trajectory fields. For single-step prediction targets, the outcome and its error appear to be the informative part of an episode, whereas trajectories may matter more for multi-step planning, which these tasks test only partly. Memory design should accordingly focus on selecting, merging and retiring verified cases rather than on recording longer traces, for example by merging repeated errors into compact lessons, preserving rare counterexamples and removing episodes that no longer improve prediction.

Experience also transferred across tasks, but selectively. With static evidence, episodes from a related physical property reduced error by 13.0--54.6\% relative to an unrelated control source, and for formation energy and both band gaps the related source even outperformed same-task memory. Unrelated memory was not neutral: for bulk modulus and the optB88 band gap, it raised the error of MatRecall above that of the same controller without episodes. Without static evidence, transfer was erratic, and cross-schema pairs gave little benefit. Agents that reuse self-reflections or extracted insights have mainly drawn on experience from the same task or environment \citep{Shinn2023Reflexion,Zhao2024ExpeL}. An LLM that shares memory across heterogeneous materials tasks instead needs an explicit decision about which experience to retrieve. A learned memory router that scores task schema, chemical similarity, property relations, uncertainty and recent episode utility could select static records, verified cases, full episodes or no memory for each query.

The comparison with specialized models suggests a division of labour. MatRecall outperformed them mainly on synthesis, nanomaterial and process tasks, in which textual or categorical context and comparable past cases are informative. On the five crystal-property tasks, it closed 49--99\% of the gap from the memory-free model but remained less accurate, and geometry-aware models were also stronger for coordinate-dominated targets in the MatText benchmark \citep{Alampara2026MatText}. A hybrid framework could therefore use the LLM to interpret tasks, manage experience and coordinate tools, while graph networks and physical simulators perform geometry-sensitive inference \citep{Xie2018CGCNN,Choudhary2021ALIGNN}. Their predictions and the eventual outcomes would then enter memory as further episodes.

Several limitations define the scope of these conclusions. All experiments use one base LLM and retrospective public datasets with immediate and complete feedback, whereas laboratory outcomes can be delayed, noisy or partial. Campaigns contain at most 48 queries per task, which is too few to expose obsolete episodes, distribution shift or forgetting. Apart from the five crystal-property tasks, the specialized models are conventional supervised models fitted to each task's training data rather than the strongest published method for each task. After feedback stopped, MatRecall outperformed static RAG on 9 of 14 tasks, so the benefit of retained experience beyond static evidence depends on the task. Finally, the comparisons are descriptive.

Evaluating several LLMs, learned routers and geometry tools on MatMemBench, and extending it to prospective experimental feedback, can establish when experience should be retrieved, shared or retired. More broadly, our results suggest that the verified outcomes of calculations and experiments, which are usually discarded once a prediction has been scored, can be accumulated as a reusable scientific resource alongside data and models. Such experience is a step towards discovery loops that become more reliable with each result they receive.

\section{Methods}

\subsection{MatMemBench construction and task metrics}

We assembled MatMemBench from public materials datasets that support task-specific prediction and evaluation (Fig.~\ref{fig:benchmark}a). Each record contains the material or process input, the required prediction schema and a separately stored outcome. The tasks cover numerical, categorical, set-valued, sequential, graph and distribution outputs. Training records populate static domain memory, whereas held-out records supply prediction queries. Source-specific metrics and units are retained when task records are converted to this common prediction--outcome format.

MatMemBench contains 22 tasks drawn from 13 public source datasets. Five tasks use JARVIS-DFT structures and the targets \texttt{bulk\_modulus\_kv}, \texttt{formation\_energy\_peratom}, \texttt{magmom\_oszicar}, \texttt{mbj\_bandgap} and \texttt{optb88vdw\_bandgap} \citep{Choudhary2020JARVIS}. The space-group task uses the Materials Project-derived SG\_predict cubic subset, and the phase task uses the released high-entropy-alloy phase labels \citep{Jain2013MP,Li2021SpaceGroup,Peivaste2023HEAPhase}. Six synthesis tasks use PcMSP for sentence-level action graphs \citep{Yang2022PcMSP}, the SyntMTE release for calcination and sintering variables \citep{Prein2025DataAugmented}, TMSR/SPENDE for operation sequences and conditions \citep{Na2023SPENDE}, SynthesisSimilarity for precursor sets \citep{He2023Precursor,He2023SynthesisSimilarityData}, the Fe--Co--Ni release for property-conditioned formulation \citep{Padhy2024InverseDesign}, and ZeoSyn for conditional synthesis-route distributions \citep{Pan2024ZeoSyn,Pan2026DiffSyn}. The InP synthesis dataset supplies observed absorption, emission and particle-size labels \citep{Nguyen2022InP}. MeltpoolNet supplies depth, length, width and the released categorical shape label \citep{Akbari2022MeltpoolNet}. The remaining process tasks use the solid-state product data of Malik and colleagues, derived from the text-mined synthesis corpus of Kononova and colleagues, and the reported and failed organically templated vanadium selenite crystallizations of Raccuglia and colleagues \citep{Kononova2019Recipes,Malik2021SynthesisOutcomes,Raccuglia2016Failed}. Each task is scored in its native metric; MAE and Wasserstein distance are minimized, while accuracy, F1, Jaccard and exact-match metrics are maximized. Supplementary Table~S1 maps every task to its source and reports the online-adaptation and memory-retention sample sizes.

Each online-adaptation task forms a separate sequence sampled from its held-out test records (Fig.~\ref{fig:benchmark}b). We use sampling seed 0 for the primary sequence. Cases are ranked by a SHA-256 hash of a fixed salt, the seed, the task name and the case identifier. For classification tasks, one case per observed class is selected first when the budget permits; the remaining places follow the hash ranking. Selected cases are then ordered by a second hash with a separate ordering namespace. This produces the same task-local order for every method, without sorting by target values. The primary cohort contains at most 48 cases per task and 942 cases overall. The separate ablation cohort contains at most 32 cases per task and 654 cases overall.

To construct the retention cohort, we select new queries whose source-native material, chemical-system, publication, topology or process groups are absent from the online-adaptation cohort. Fourteen tasks contribute at least ten such queries each, yielding 429 queries evaluated with fixed same-task memory (Fig.~\ref{fig:benchmark}c). An inverse-formulation extension uses evidence-anchor identity as its grouping unit and is reported separately in the Supplementary Information. Cross-task transfer draws target queries from the retention cohorts and substitutes fixed memories acquired on relevant or unrelated source tasks (Fig.~\ref{fig:benchmark}d). Thus, retention and transfer reuse online experience with different memory origins, while both withhold target feedback.

\subsection{Static multiview retrieval}

Static domain memory contains training input--output examples (Fig.~\ref{fig:overview}d). Retrieval views are selected by task family. Composition views compare reduced formula, element sets and stoichiometric fractions. Structure views use crystal descriptors and structural-neighbour information when coordinates are available. Process views compare precursors, operations, conditions and textual fields. Candidate ranks from the available views are fused before constructing the evidence context. Static RAG uses this context without writing new cases. ReAct + static RAG \citep{Yao2023ReAct} exposes the same retrievers as tools and allows the LLM to decide whether to retrieve again, but the tool state remains fixed across queries.

\subsection{Verified episodic memory}

The four-stage cycle in Fig.~\ref{fig:overview}d defines a prediction episode as the span from receiving a query to verifying its outcome. MatRecall represents each completed episode as
\begin{equation}
e_t=(q_t, a_t, z_t, \hat{y}_t, y_t, \ell_t, m_t),
\end{equation}
where $q_t$ is the query, $a_t$ is the action and tool trajectory, $z_t$ is the retrieved evidence, $\hat{y}_t$ is the prediction, $y_t$ is the verified outcome, $\ell_t$ is the task-native error or score and $m_t$ is a compressed lesson. The lesson is derived by rule from the other fields and records the direction of the prediction error, the additional static view requested by the controller, the score change relative to a deterministic estimate from the retrieved static neighbours and whether the model returned a valid response. Outcome verification defines the encoding boundary. The episode is written only after $\hat{y}_t$ has been serialized and scored. Retrieval for query $q_t$ is restricted to static records and episodes with index smaller than $t$. The current outcome is therefore unavailable at prediction time.

Episode retrieval uses the same task-specific query views as static retrieval, augmented with error direction, prediction schema, evidence type and lesson text. The selected records are reinstated in the LLM's working context for explicit reasoning. Dynamic case RAG writes only $(q_t,y_t)$ and retrieves these verified pairs. Rolling-context ICL places the most recent $k$ verified pairs directly in the prompt without a persistent search index. MatRecall retrieves verified cases and full episodes; the ReAct controller may request an additional static view, inspect a past error or stop and predict. The episode writer stores both the original structured fields and a token-bounded summary.

\subsection{Online adaptation and ablation protocols}

The online-adaptation setting of MatMemBench evaluates 942 sequentially presented cases across the 22 tasks, with at most 48 cases per task. Each of the six LLM methods receives the same ordered queries and reveals each target only after all required outputs have been recorded. The specialized models predict the same query identifiers without online updates. The primary result is the aggregate native metric over the complete sequence, not only the final query. Every method is therefore compared on the same cohort using the task's native metric. The memory acquired over this sequence is the memory evaluated in the retention experiments.

The ablation study uses a separate cohort with at most 32 cases per task and 654 cases per sequence. It includes five shuffled sequences and five grouped sequences. Grouped sequences place cases from the same source-defined material or process group next to one another. The memory and feedback ablations are ReAct + verified cases, prediction + outcome, full trajectory, score only and delayed write, each compared with MatRecall on the same queries. Retrieval variants retrieve $k=2,4,6,8,10$ or 12 episodes per query. The order analysis contains 220 paired task--sequence comparisons from 22 tasks and ten sequences. The experience analysis compares the first and later members of repeated groups over the same 15 tasks.

\subsection{Memory retention after feedback stops}

In the memory-retention setting of MatMemBench, the retained memory is the snapshot produced by the completed online-adaptation sequence. Retention queries are group-disjoint from online-adaptation queries and contain no repeated evaluation record. The memory snapshot is identical for every query within a method. No target is revealed and no episode is written during evaluation. The factorial ablation compares four memory conditions that cross static evidence with verified cases or full episodes: cases only, dynamic case RAG, episodes only and MatRecall. LLM-only, static RAG, ReAct + static RAG and specialized models are evaluated on the same retention queries.

For tasks where the specialized model improves on LLM-only, the fraction of the gap closed by MatRecall is
\begin{equation}
g=\frac{s_{\mathrm{MatRecall}}-s_{\mathrm{LLM}}}{s_{\mathrm{specialized}}-s_{\mathrm{LLM}}},
\end{equation}
after orienting every metric so that larger values are better. Values above one indicate that MatRecall reaches or exceeds the specialized-model score on that cohort.

\subsection{Cross-task transfer}

The cross-task transfer setting of MatMemBench keeps target queries fixed and changes only the source-task memory. Each target has one relevant source task and one unrelated control task selected before scoring. The primary set contains four schema-matched regression targets. The additional set contains four phase, symmetry or synthesis targets. All source memories remain fixed, and target answers are not revealed. Each comparison uses the same target queries and reports the task-native performance change between relevant and control source memory.

\subsection{LLM and specialized baselines}

All LLM methods use the same Qwen/Qwen3.8-27B checkpoint, task input, response schema and decoding configuration within an experiment \citep{QwenTeam2026Qwen38}. Prompts instruct the model to return a structured prediction and preserve the requested output schema when retrieved records contain a different representation. For the five crystal-property tasks, the specialized models are ALIGNN 2.0 checkpoints matched to the five JARVIS targets; the JARVIS data, original ALIGNN architecture and ALIGNN 2.0 implementation are cited separately \citep{Choudhary2020JARVIS,Choudhary2021ALIGNN,Lee2026ALIGNN2}. For the other 17 tasks, the specialized models are supervised models fitted on each task's training partition and selected on validation data. They comprise random forests \citep{Breiman2001RandomForests}, extremely randomized trees \citep{Geurts2006ExtraTrees}, nearest-neighbour models with cosine or TF--IDF similarity \citep{Cover1967NearestNeighbor,Salton1988TFIDF}, two-stage classification--regression models and token--relation logistic regression \citep{Cox1958BinarySequences}, implemented with scikit-learn \citep{Pedregosa2011ScikitLearn}. Supplementary Table~S2 specifies the task-level model family and selection candidates. Specialized predictions are evaluated independently and never enter static or episodic memory.

\subsection{Statistical analysis}

The task is the primary unit for comparisons because metrics and units differ. We report complete task-native values for online adaptation, memory retention and cross-task transfer. For ablation summaries, each paired change is expressed as a percentage of the baseline value. Lower-is-better metrics are reversed so that positive values always indicate improvement. We report the median percentage across tasks. All summaries are descriptive; no hypothesis tests or P values are reported.

\section*{Data availability}

All source datasets are public releases. Crystal-property, space-group and phase tasks use JARVIS-DFT, the Materials Project-derived SG\_predict release and ML\_HEAs\_Phase\_Dataset, respectively \citep{Choudhary2020JARVIS,Jain2013MP,Li2021SpaceGroup,Peivaste2023HEAPhase}. Synthesis tasks use PcMSP, SyntMTE, TMSR/SPENDE, the SynthesisSimilarity Dryad release, the Fe--Co--Ni inverse-design release and ZeoSyn \citep{Yang2022PcMSP,Prein2025DataAugmented,Na2023SPENDE,He2023Precursor,He2023SynthesisSimilarityData,Padhy2024InverseDesign,Pan2024ZeoSyn,Pan2026DiffSyn}. Nanomaterial and process tasks use the InP quantum-dot synthesis dataset, MeltpoolNet, the Kononova--Malik solid-state synthesis corpus and the Raccuglia crystallization dataset \citep{Nguyen2022InP,Akbari2022MeltpoolNet,Kononova2019Recipes,Malik2021SynthesisOutcomes,Raccuglia2016Failed}. MatMemBench, comprising the processed online-adaptation, memory-retention and cross-task transfer data, and the processed MatText benchmark data are available on GitHub at \url{https://github.com/ZHymLumine/MatRecall/tree/main/data}; the original datasets remain under their source terms. Numerical source data underlying the figures and tables are provided with the Supplementary Information.

\section*{Code availability}

The code for MatRecall is released under the MIT licence and is available on GitHub at \url{https://github.com/ZHymLumine/MatRecall}.

\section*{Acknowledgements}

This work is supported by JST ARISE (JPMJAR2601 and JPMJAR2605).

\section*{Author contributions}

Y.Z. conceived and implemented the study, curated the benchmarks, performed the experiments and drafted the manuscript. K.T. supervised the study and revised the manuscript. Both authors interpreted the results and approved the final manuscript.

\section*{Competing interests}

The authors declare no competing interests.

\begingroup
\setlength{\bibsep}{3pt plus 1pt minus 1pt}
\bibliography{references}
\endgroup

\clearpage
\pdfximage{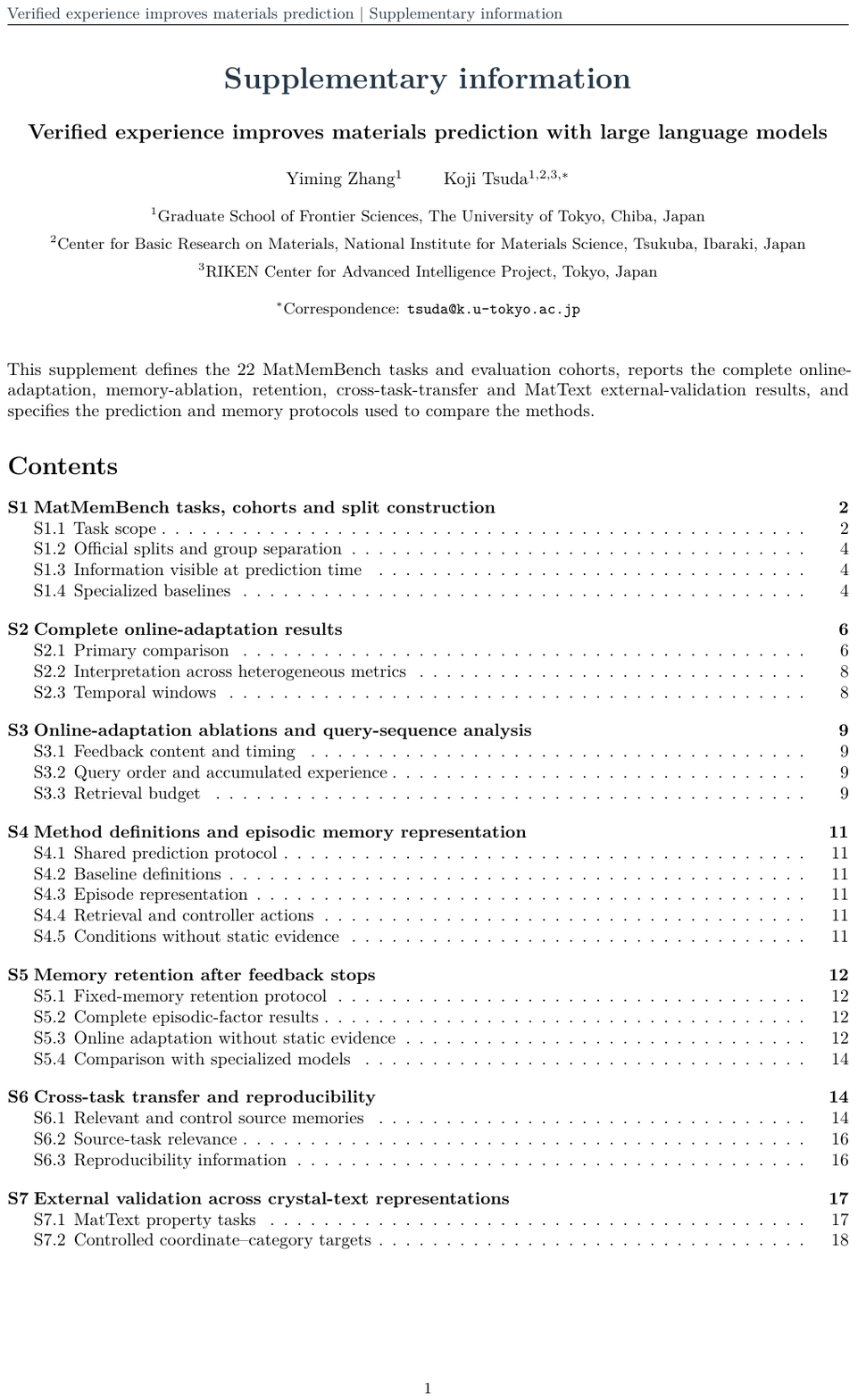}
\edef\supplementpagecount{\the\pdflastximagepages}
\newcount\supplementpage
\supplementpage=1
\loop
  \includepdf[pages=\the\supplementpage,fitpaper=true]{supplementary.pdf}
  \advance\supplementpage by 1
\ifnum\supplementpage<\numexpr\supplementpagecount+1\relax
\repeat

\end{document}